\documentclass[10pt,twocolumn,letterpaper]{article}
\pdfoutput=1
\usepackage{iccv}
\usepackage{times}
\usepackage{epsfig}
\usepackage{graphicx}
\usepackage{amsmath}
\usepackage{amssymb}
\usepackage[marginal]{footmisc}

\usepackage[breaklinks=true,bookmarks=false]{hyperref}

\iccvfinalcopy 

\def\iccvPaperID{****} 

\ificcvfinal\fi

\begin{document}

\title{Sequential Adversarial Learning for Self-Supervised Deep Visual Odometry}

\author{Shunkai Li \qquad
	    Fei Xue\thanks{equal contribution} \qquad
	    Xin Wang\textsuperscript{*} \qquad
	    Zike Yan\qquad
	    Hongbin Zha\\
Key Laboratory of Machine Perception (MOE), School of EECS, Peking University\\
PKU-SenseTime Machine Vision Joint Lab\\
{\tt\small \{lishunkai, feixue, xinwang\_cis, zike.yan\}@pku.edu.cn \quad zha@cis.pku.edu.cn} 
}

\maketitle
\ificcvfinal\thispagestyle{empty}\fi

\begin{abstract}
   We propose a self-supervised learning framework for visual odometry (VO) that incorporates correlation of consecutive frames and takes advantage of adversarial learning. Previous methods tackle self-supervised VO as a local structure from motion (SfM) problem that recovers depth from single image and relative poses from image pairs by minimizing photometric loss between warped and captured images. As single-view depth estimation is an ill-posed problem, and photometric loss is incapable of discriminating distortion artifacts of warped images, the estimated depth is vague and pose is inaccurate. In contrast to previous methods, our framework learns a compact representation of frame-to-frame correlation, which is updated by incorporating sequential information. The updated representation is used for depth estimation. Besides, we tackle VO as a self-supervised image generation task and take advantage of Generative Adversarial Networks (GAN). The generator learns to estimate depth and pose to generate a warped target image. The discriminator evaluates the quality of generated image with high-level structural perception that overcomes the problem of pixel-wise loss in previous methods. Experiments on KITTI and Cityscapes datasets show that our method obtains more accurate depth with details preserved and predicted pose outperforms state-of-the-art self-supervised methods significantly.
\end{abstract}

\section{Introduction}
The ability for an agent to understand 3D environment and infer ego-motion is crucial for many real-world applications, such as autonomous driving~\cite{driving}, robotics~\cite{robot}, and virtual/augmented reality~\cite{VR}. As the problem of simultaneous localization and mapping (SLAM) and visual odometry (VO) has a clear meaning in 3D geometry, VO/SLAM has been studied as a multi-view geometric problem for decades. These classic methods~\cite{DSO,LSD,svo,PTAM,orb} perform well in regular scenes, but fail in challenging conditions due to their inherent reliance on low-level feature correspondences. Since deep learning captures structural perception by extracting high-level features, a number of learning-based VO methods have been applied to break through the limitations of classic approaches~\cite{mapnet,demon,deepvo,guided,beyond,deeptam}.

\begin{figure}
	\begin{center}
		\includegraphics[width=1.0\linewidth]{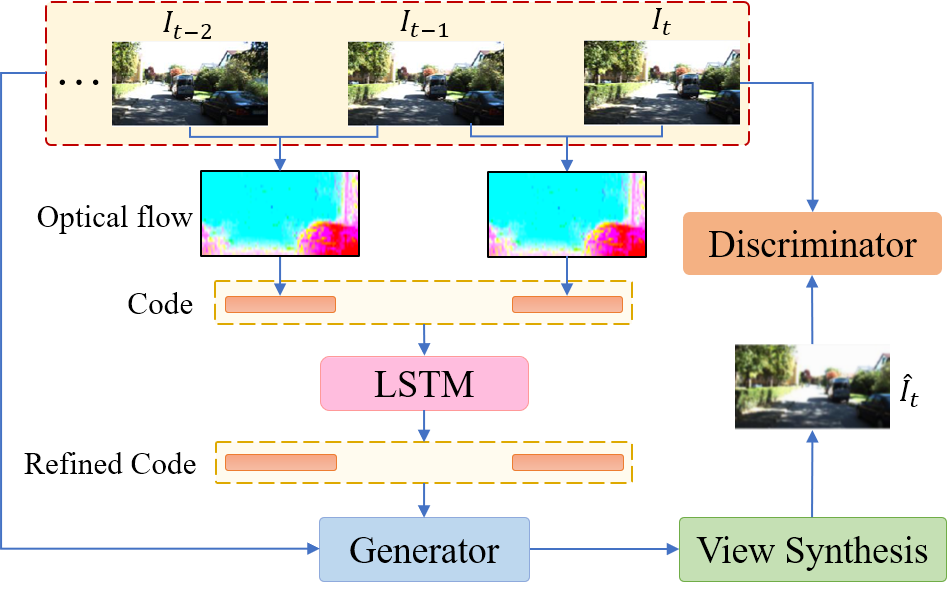}
	\end{center}
	\caption{Overview of our method. The network extracts optical flow into a compact code, which is incorporated by LSTM to aggregate historical information and refine previous estimations. Depth and pose estimation is regarded as an image conditioned generative task, and the refined code is provided as input signal. The geometric inference is used to reconstruct a warped image by view synthesis and evaluated by a discriminator.}
	\label{fig:overview}
\end{figure}

However, supervised learning requires substantial labeled data, which is either tedious or impractical to obtain. Recent work has been trying to address this problem by coupling depth and pose estimation in a self-supervised manner~\cite{GeoNet,SfMLearner}. As image sequence is the only input, all the estimations should be mapped to image space for self-supervision. The mapping is typically made by view synthesis and photometric loss is defined to minimize the difference between synthesized image and the real one.

In self-supervised VO, estimation of depth and pose are simultaneously learned in a coupled way, accurate depth contributes to precise pose estimation and vice versa. Previous works on self-supervised VO estimate depth from single view. As an ill-posed problem, the output depth is vague, hence the predicted pose is also inaccurate. However, uncertainties in depth estimation can be eliminated by exploiting correlations between consecutive frames. Nonetheless, due to the data redundancy of image sequence, it is inefficient to integrate the information of multiple frames by stacking them along the RGB channel. In this paper, we propose to learn a compact representation (referred to as `code') of the correlation between frames, and sequential information is accumulated by integrating codes via Long Short-Term Memory (LSTM). The code provides correlations of consecutive frames that help generate clear depth maps and reduce accumulated error over a long sequence. 

On the other hand, inaccurate depth and pose leads to distortion artifacts in synthesized images (Fig.~\ref{fig:warp}), which are difficult to be eliminated by photometric loss due to the pixel-level correspondence. A new evaluation criterion with structural perception is needed for accurate depth estimation. In this paper, we tackle VO as a self-supervised image generation task and take advantage of Generative Adversarial Networks (GAN)~\cite{GAN}. The generator learns to estimate depth and pose to synthesize a warped image, while the discriminator evaluates the quality of synthesized image with structural perception and higher-level understandings. This two-player game impels the generator to estimate more accurate depth and pose, while the discriminator is able to distinguish distortion artifacts with structural perception.

The overview of our method is shown in Fig.~\ref{fig:overview}. Different from single-view estimation, our method generates clear depth with additional information which cannot be retrieved from a single image. The information is obtained by encoding optical flow into a compact code, and codes of multiple frames are incorporated and refined  by LSTM. The overall framework is treated as a generative model with adversarial learning. During training, the spatial-temporal consistency is enforced as self-supervision. The main contributions of our paper can be summarized as follows:
\begin{itemize}
	\item We propose to exploit spatial-temporal correlations over long sequence to significantly reduce estimation errors and scale drift for self-supervised VO.
	\item We treat self-supervised VO as a generative model and take the advantage of adversarial learning for self-supervised pose and depth estimation.
\end{itemize}

Our method outperforms state-of-the-art self-supervised approaches significantly and gives comparable results with supervised manners. Extensive experiments manifest the advantages of our model. Besides, the idea of self-supervised adversarial learning with spatial-temporal consistency may also bring insight into VO/SLAM and video-based computer vision researches.

\section{Related works}
Humans are capable of perceiving 3D environment and inferring ego-motion in a short time, but it is hard for an agent to be equipped with similar capabilities. VO/SLAM has been considered as a multi-view geometric problem for decades. It is traditionally solved by minimizing photometric~\cite{LSD} or geometric~\cite{orb} reprojection errors and works well in regular environments, but fails in challenging conditions like dynamic objects and abrupt motions. In light of these limitations, VO has been studied with learning techniques in recent years and many approaches with promising performance have been proposed.

{\bf Supervised methods} formulate VO as a supervised learning problem and many methods with good results have been proposed. DeMoN~\cite{demon} jointly estimates pose and depth in an end-to-end manner. Inspired by the practice of parallel tracking and mapping in classic VO/SLAM, DeepTAM~\cite{deeptam} utilizes two networks for pose and depth estimation. DeepVO~\cite{deepvo} treats VO as a sequence-to-sequence learning problem by estimating poses recurrently. The limitation of supervised learning is that it requires a large amount of labeled data. The acquisition of ground truth often requires expensive equipment or highly manual labeling, and some gathered data are inaccurate. Depth obtained by LIDAR is sparse, and the output depth of Kinect contains a lot of noise. Furthermore, some ground truth is unable to obtain (\eg optical flow). Previous works have tried to address these problems with synthetic datasets~\cite{flownet}, but there is always a gap between synthetic and real-world data.

{\bf Self-supervised methods} In order to alleviate the reliance on ground truth, recently many self-supervised methods have been proposed for VO. The key to self-supervised learning is to find the internal correlations and constraints in the training data. SfMLearner~\cite{SfMLearner} leverages the geometric correlation of depth and pose to learn both of them in a coupled way, with a learned mask to mask out regions that don't meet static scene assumption. As the first self-supervised approach for VO, SfMLearner couples depth and pose estimations with image warping, which becomes the problem of minimizing photometric loss. Inherited from this idea, many self-supervised VO have been proposed, including modifications on loss functions~\cite{CTC,undeepvo}, network architectures~\cite{deeperundemon,undemon,CTC,vid2depth,deepvofeat}, predicted contents~\cite{GeoNet}, and combination with classic VO/SLAM~\cite{driven,DVSO}. For example, GeoNet~\cite{GeoNet} extends the framework to jointly estimate optical flow with forward-backward consistency to infer unstable regions and achieves state-of-the-art performance among self-supervised VO methods.

Despite its feasibility, self-supervised VO still underperforms supervised ones. Apart from the effectiveness of direct supervision, a key reason is that they focus mainly on geometric properties~\cite{SfMLearner} but pay little attention to the sequential nature of the problem. In these methods, only a few frames (no more than 5) are processed in the network, while previous estimations are discarded and the current estimation is made from scratch. Instead, the performance can be enhanced by taking geometric relations of sequential observations into account.

Our approach differs from previous art in formulating self-supervised VO as a sequential learning problem. The frame-to-frame correlation is represented as a compact code, and sequential information are integrated via LSTM. In contrast to prevalent single-view depth estimation, our framework estimates depth with the code conditioned on a single image and treat VO as a generative task. By means of adversarial learning, our method provides sharper depth and more accurate pose estimations.

\section{Method}
In this section, we will introduce our method in detail. The entire framework consists of four components (Fig.~\ref{framework}). The encoder extracts high-level features from optical flow into a compact code in Sec.~\ref{encoder}, and the codes are aggregated and further refined by LSTM in Sec.~\ref{refinement}. The generator estimates depth and pose conditioned on refined codes and images in Sec.~\ref{depth}-\ref{poseandmask}. The discriminator in Sec.~\ref{discriminator} judges the authenticity of a synthesized view. Finally, loss functions used in training are defined in Sec.~\ref{loss}.

\subsection{Encoder}\label{encoder}
Visual odometry estimates camera motion between consecutive image pairs. This estimation is computed by feature correspondence or photometric consistency in classic VO/SLAM. Different from previous self-supervised methods that estimate directly from raw images, we provide the network with a representation of frame-to-frame correspondence for depth and pose estimation.

As a way of frame-to-frame correspondence, parallax and motion of each pixel can be obtained by computing optical flow between consecutive images. In our framework, we compute optical flow~\cite{classicopticalflow} and extract it into a compact representation (referred to as `code') $c_{t}$ with a size of 128
\begin{equation}
c_{t}=\mathcal{C}(\mathcal{F}(I_{t-1}, I_{t})).
\end{equation}
The extracted $c_{t}$ will be incorporated with historical information and used as side input for depth and pose estimation.

\subsection{Sequential information aggregation}\label{refinement}
Estimating depth and pose from only a few frames is prone to error accumulation and scale drift. The problem can be mitigated by exploiting correlations over long sequence. This formulation is appealing for self-supervised sequential estimations since it utilizes incoming observations and spatial-temporal consistency as self-supervision.

In our framework, we use LSTM~\cite{LSTM} to model VO as a self-supervised sequential learning problem. As an extension of recurrent neural networks (RNN), LSTM introduces a \emph{cell} to remember and forget information adaptively. LSTM fuses the code $c_{t}$ of current frame $I_{t}$ into accumulated information. Intuitively, the long-term information is remembered as a prior, and short-term memory is used to infer the current state. The feature flow passing through recurrent units carries rich information of previous states, enabling refined outputs to improve the current estimation
\begin{equation}
c^{'}_{t}, h_{t}=\mathcal{U}(c_{t},h_{t-1}),
\end{equation}
where $c^{'}_{t}$ denotes the refined code that incorporates historical information, and $h_{t-1}$, $h_{t}$ are hidden states at time $t-1$, $t$, respectively.

\begin{figure*}
	\begin{center}
		\includegraphics[width=1\linewidth]{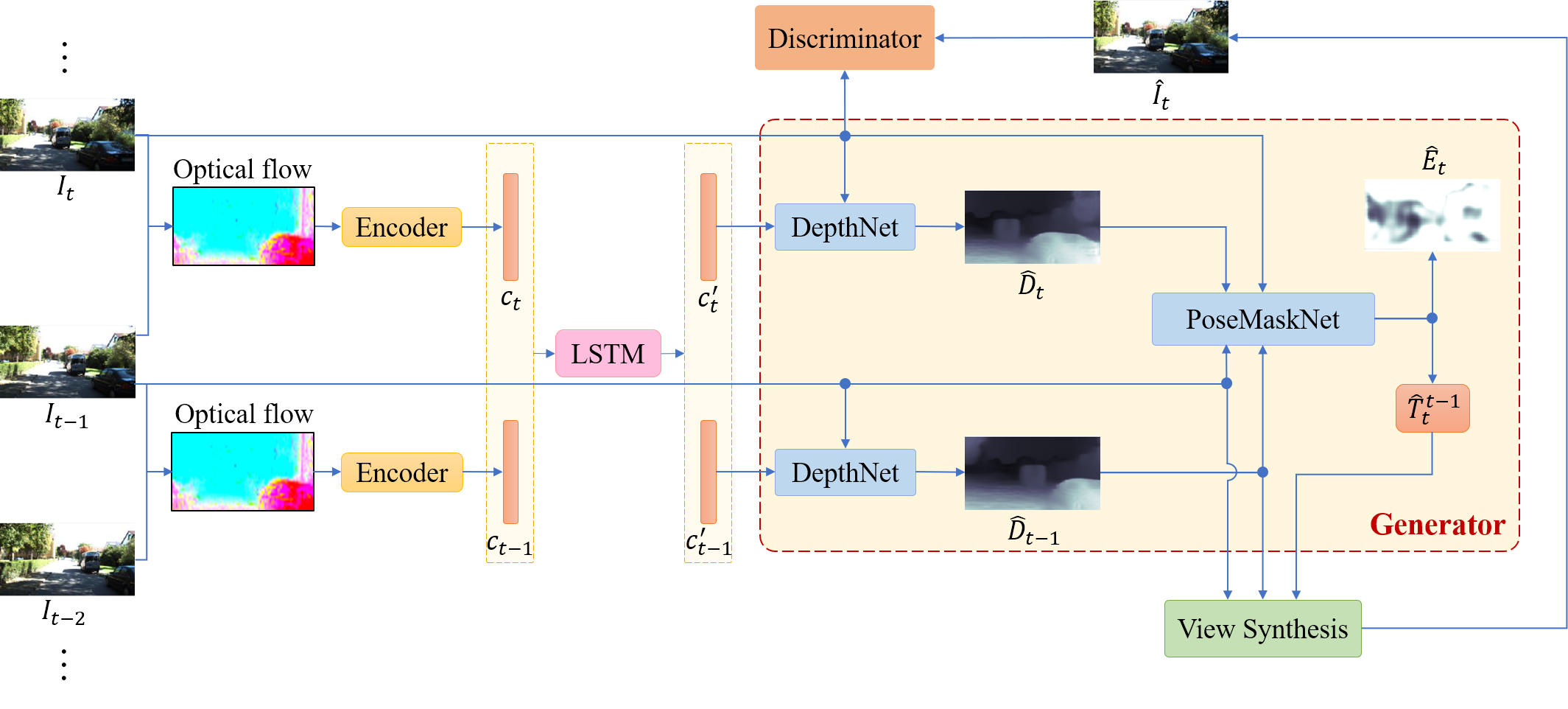}
	\end{center}
	\caption{Illustration of our framework. The encoder compresses optical flow of two consecutive images into a compact code, which is aggregated and refined by LSTM. The DepthNet estimates depth conditioned on the refined code and input image. The estimated depth is concatenated with image for pose and mask prediction, while the authenticity of the warped image is judged by the discriminator. The discriminator is excluded during the test phase.}
	\label{framework}
\end{figure*}

\subsection{Depth estimation}\label{depth}
In the existing literature, depth is estimated from a single image \emph{I}
\begin{equation}
\hat{D}=\mathcal{D}(I).
\end{equation}
As an ill-posed problem, the estimated depth is reasonable on the whole but vague in details. On the other hand, simply stacking multiple frames does not improve the result of depth estimation~\cite{SfMLearner}. In order to obtain a clear depth, correlations of multiple views should be provided as additional information which cannot be retrieved from a single image.

Because of the high degree of order and regularity of 3D scenes, depth can be effectively represented by a compact feature with a single image~\cite{codeslam}. As motion parallax of two frames reflects the distance of each part of the scene, we provide the refined code $c^{'}_{t}$ as side input for depth estimation
\begin{equation}
\hat{D}_{t}=\mathcal{D}(I_{t}, c^{'}_{t}).
\end{equation}

As an image conditioned depth generation process, $I_{t}$ is extracted into a feature map by convolutional layers, which is further concatenated with $c^{'}_{t}$ in the network. It is then followed by up-sampling layers with skip connections.

\subsection{Pose and mask estimation}\label{poseandmask}
Most self-supervised VO methods regress pose directly from images but fail to exploit the depth of two views. In classic methods, pose regression from images and depth is solved by RGBD registration, such as using image feature detection for initial guess and robust 3D correspondence for pose refinement~\cite{RGBDreg,RGBDreg2}. In order to exploit both color and depth information, we stack images and depth maps into 2 RGBD images for pose estimation from $t-1$ to $t$
\begin{equation}
\hat{T}^{t}_{t-1}=\mathcal{P}((I_{t-1}, \hat{D}_{t-1}), (I_{t}, \hat{D}_{t})).
\label{equation_pose}
\end{equation}

After the acquisition of pose and depth, image warping is used for view synthesis. The homogeneous coordinate of a pixel in the target view $p_{t}$ and the source view $p_{t-1}$ are correlated by~\cite{SfMLearner}
\begin{equation}
p_{t-1} \sim K\hat{T}^{t-1}_{t}\hat{D}_{t}(p_{t})K^{-1}p_{t},
\label{warping}
\end{equation}
where $K$ denotes camera intrinsics. We use differentiable bilinear sampling as~\cite{SfMLearner}. In this way, the synthesized image $\hat{I}_{t}$ and $I_{t}$ can be used for self-supervision.

Nonetheless, view synthesis builds on the assumption that the scene is static without illumination change and occlusions, which is often violated in practice. To overcome this problem, our framework learns to predict a per-pixel mask $\hat{M}_{t}$ as a belief in how successful a target pixel is rendered during view synthesis~\cite{SfMLearner}. Consequently, the weighted photometric loss is
\begin{equation}
\mathcal{L}_{pho}=\sum_{<I_{1},...,I_{N}>}\sum_{p}\hat{M}_{t}(p)\|\hat{I}_{t}(p)-I_{t}(p)\|_{1}.
\label{photoloss}
\end{equation}

\subsection{Discriminator}\label{discriminator}
\begin{figure}
	\begin{center}
		\includegraphics[width=1.0\linewidth]{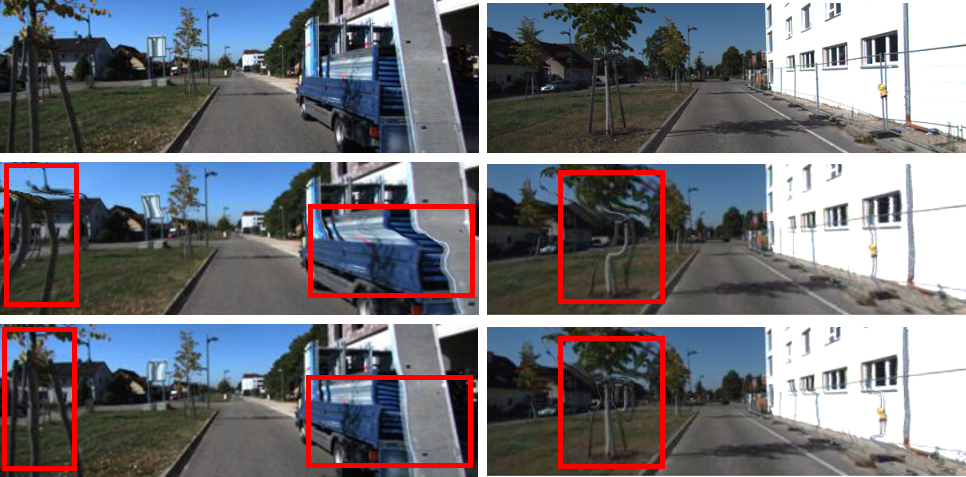}
	\end{center}
	\caption{Example of warped images according to the estimated depth and pose. Top row: captured images, medium row: warped images of SfMLearner~\cite{SfMLearner}, bottom row: warped images of our method. It can be seen that inaccurate predictions will lead to distortion artifacts on the warped image. Compared to the existing literature, our method synthesizes more accurate warped images.}
	\label{fig:warp}
\end{figure}

Photometric loss is widely used in self-supervised VO and the warped results are shown in Fig.~\ref{fig:warp}. Despite convolutional neural networks (CNN) extract high-level features that prevent low-level feature problem in classic VO/SLAM, the loss function is still based on pixel-level instead of evaluating on a larger receptive field with higher-level understandings. Due to the pixel-level correspondence and photometric consistency assumption, photometric loss is not robust to occlusion, texture-less regions, dynamic objects and illumination change. In these challenging conditions, there are multiple local minima with similar magnitudes. The network tends to trap into any of them during training with vague depth and wrong pose, leading to inaccurate reconstruction (Fig.~\ref{fig:warp}). Some of previous research have realized this problem~\cite{GeoNet,deepvofeat} and try to eliminate this disturbing factor by explicitly modeling motion segmentation and optical flow, but achieve limited improvement.

Instead, the distortion artifacts are easily detectable by a discriminator. The compelling results achieved by GAN have been successfully demonstrated in many image generation tasks~\cite{GANdepth,pix2pix,cyclegan}. The adversarial learning impels the network to learn more flexible distributions to tackle underfitting issues and overcome gradient locality. In the self-supervised paradigm, VO can be regarded as a conditional image generation task
\begin{equation}
\hat{I_{t}}=G(c^{'}_{t-1}, c^{'}_{t}|I_{t-1}, I_{t}).
\end{equation}
$I_{t}$ is a sample from distribution $p_{real}$, and $\hat{I}_{t}$ is generated from $c^{'}_{t-1}$, $c^{'}_{t}$ on the latent space $p_{code}$. 

During training, the generator tries to fool the discriminator by generating better pose and depth. Meanwhile, given $I_{t}$ as side information, the discriminator tries to distinguish the fake $\hat{I_{t}}$ by predicting a probability of authenticity $D(\hat{I_{t}}|I_{t})$. The adversarial training overcomes the problem of Eq.~\eqref{photoloss} to produce accurate depth and pose without explicit modeling of motion segmentations and optical flow. The value function of this min-max game can be formulated according to~\cite{pix2pix}
\begin{equation}
\begin{aligned}
\mathcal{L}_{GAN}=&\min\limits_{G}\max\limits_{D}V(G,D) \\
=&\mathbb{E}_{I_{t}\sim p_{real}}[\log(D(I_{t}|I_{t}))]+ \\
&\mathbb{E}_{c^{'}_{t-1},c^{'}_{t}\sim p_{code}}[\log(1-D(\hat{I_{t}}|I_{t}))].
\end{aligned}
\label{GAN}
\end{equation}

\subsection{Loss functions}\label{loss}
{\bf Appearance loss} In order to overcome the pixel-level correspondence problem, we measure the reconstructed images from both weighted photometric loss and structural similarity metric (SSIM)~\cite{ssim}
\begin{equation}
\begin{aligned}
\mathcal{L}_{ap}=&\mathcal{L}_{reg}(\hat{M})+(1-\alpha)\mathcal{L}_{pho} \\
&+\frac{1}{N}\sum_{x,y}\alpha\frac{SSIM(\hat{I}(x,y), I(x,y))}{2},
\end{aligned}
\end{equation}
$\mathcal{L}_{reg}(\hat{M})$ is a regularization term to prevent the network converges to a trivial solution, which is detailed in~\cite{SfMLearner}. $N$ is the number of images in the training minibatch. The filter size of SSIM is set 10$\times$10 and $\alpha$ is set 0.85.

{\bf Depth regularization} Discontinuity of depth usually happens where strong image gradients are present. Similar to~\cite{undemon,deepvofeat}, we introduce an edge-aware smoothness loss to enforce discontinuity and local smoothness in depth
\begin{equation}
\begin{aligned}
\mathcal{L}_{smo}=&\frac{1}{N}\sum_{x,y}\|\nabla_{x}\hat{D}(x,y)\|e^{-\|\nabla_{x}I(x,y)\|}+ \\&\|\nabla_{y}\hat{D}(x,y)\|e^{-\|\nabla_{y}I(x,y)\|}.
\end{aligned}
\end{equation}

{\bf Trajectory consistency}
Although LSTM-based framework is suffice to provide more accurate poses by filtering out the noise between consecutive transformations, the estimated $\hat{T}^{t-1}_{t}$ are still relative poses. There are no relations and no geometric consistency among them. Actually, these relative poses can be transformed into a unified coordinate by accumulating them along the trajectory. According to rigid-body transformation, given a set of transformations such as $A\to B\to C\to D$, the relative poses $T^{B}_{A}$, $T^{C}_{B}$, $T^{D}_{C}$ satisfies the following constraints~\cite{CTC}
\begin{equation}
\begin{aligned}
T^{B}_{A}\cdot T^{C}_{B}\cdot T^{D}_{C}&=T^{D}_{A}, \\
T^{B}_{A}\cdot T^{C}_{B}&=T^{C}_{A}, \\
T^{C}_{B}\cdot T^{D}_{C}&=T^{D}_{B},
\end{aligned}
\end{equation}
In order to enforce trajectory consistency, we compute the following loss on three scales for every eight frames
\begin{equation}
\begin{aligned}
\mathcal{L}_{TC}=\frac{1}{N}\sum_{i=1}^{N}\sum_{t\in[2,4,8]}\|\hat{{p}_{d}}^{i+t}_{i}-\hat{{p}_{r}}^{i+t}_{i}\|_{1},
\end{aligned}
\end{equation}
where $\hat{{p}_{d}}^{i+t}_{i}$ is the 6-DoF pose directly estimated from ($I_{i}$, $c^{'}_{i}$) and ($I_{i+t}$, $c^{'}_{i+t}$), and $\hat{{p}_{r}}^{i+t}_{i}$ is the concatenated 6-DoF pose of successive relative transformations.

{\bf GAN loss} in Eq.~\eqref{GAN} acts as an auxiliary self-supervision for the synthesized image. The final loss function becomes
\begin{equation}
\mathcal{L}_{final}=\lambda_{a}\mathcal{L}_{ap}+\lambda_{s}\mathcal{L}_{smo}+\lambda_{t}\mathcal{L}_{TC}+\lambda_{g}\mathcal{L}_{GAN}.
\end{equation}

\section{Experiments}
In this section, we will introduce the implementation details and show both qualitative and quantitative results compared with other methods. In the end, an ablation study is employed to test the effectiveness of each component in our framework.

\subsection{Implementation details}
\begin{table*}
	\footnotesize
	\begin{center}
		\begin{tabular}{|l|c|c|c|c|c|c|c|c|c|c|}
			\hline
			Method             & Supervision & Dataset & Cap & Abs Rel & Sq Rel & RMSE & RMSE log & $\delta<1.25$ & $\delta<1.25^{2}$ & $\delta<1.25^{3}$ \\
			\hline
			Train set mean     & -           & K       & 80m & 0.361   & 4.826  & 8.102 & 0.377 & 0.638 & 0.804 & 0.894 \\
			\hline
			Eigen \etal~\cite{eigen} Coarse & Depth    & K       & 80m & 0.214   & 1.605  & 6.563 & 0.292 & 0.673 & 0.884 & 0.957 \\
			Eigen \etal~\cite{eigen} Fine   & Depth   & K       & 80m & 0.203   & 1.548  & 6.307 & 0.282 & 0.702 & 0.890 & 0.958 \\
			Liu \etal~\cite{liu}          & Depth    & K       & 80m & 0.201   & 1.584  & 6.471 & 0.273 & 0.680 & 0.898 & 0.967 \\
			SfMLearner~\cite{SfMLearner}& -  & K       & 80m & 0.208   & 1.768  & 6.856 & 0.283 & 0.678 & 0.885 & 0.957 \\
			Vid2Depth~\cite{vid2depth} & - & K     & 80m & 0.163   & 1.240  & 6.220 & 0.250 & 0.762 & 0.916 & 0.968 \\
			GeoNet~\cite{GeoNet} & -        & K       & 80m & 0.155   & 1.296  & 5.857 & 0.233 & 0.793 & 0.931 & 0.973 \\
			Zhan \etal~\cite{deepvofeat} & Stereo & K       & 80m & \textbf{0.135} & 1.132 & 5.585 & 0.229 & 0.820 & 0.933 & 0.971 \\
			Ours               & -           & K       & 80m & 0.150 & \textbf{1.127} & \textbf{5.564} & \textbf{0.229} & \textbf{0.823} & \textbf{0.936} & \textbf{0.974}  \\
			\hline
			Garg \etal~\cite{garg} & Stereo  & K       & 50m & 0.169   & 1.080  & 5.104 & 0.273 & 0.740 & 0.904 & 0.962 \\
			SfMLearner~\cite{SfMLearner} & - & K       & 50m & 0.201   & 1.391  & 5.181 & 0.264 & 0.696 & 0.900 & 0.966 \\
			Vid2Depth~\cite{vid2depth} & - & K      & 50m & 0.155   & 0.927  & 4.549 & 0.231 & 0.781 & 0.931 & 0.975 \\
			GeoNet~\cite{GeoNet}  & -        & K       & 50m & 0.147   & 0.936  & 4.348 & 0.218 & 0.810 & 0.941 & 0.977 \\
			Zhan \etal~\cite{deepvofeat} & Stereo & K  & 50m & \textbf{0.128} & \textbf{0.815} & 4.204 & 0.216 & \textbf{0.835} & 0.941 & 0.975 \\
			Ours               & -           & K       & 50m & 0.146 & 0.927 & \textbf{4.107} & \textbf{0.216} & 0.819 & \textbf{0.943} & \textbf{0.981}  \\
			\hline
			SfMLearner~\cite{SfMLearner} & - & CS+K    & 80m & 0.198   & 1.836  & 6.565 & 0.275 & 0.718 & 0.901 & 0.960 \\
			Vid2Depth~\cite{vid2depth} & - & CS+K & 80m & 0.159 & 1.231  & 5.912 & 0.243 & 0.784 & 0.923 & 0.970 \\
			GeoNet~\cite{GeoNet}  & -        & CS+K    & 80m & 0.153   & 1.328  & 5.737 & 0.232 & 0.802 & 0.934 & 0.972 \\
			Ours               & -           & CS+K    & 80m & \textbf{0.136} & \textbf{1.064} & \textbf{5.176} & \textbf{0.289} & \textbf{0.830} & \textbf{0.942} & \textbf{0.976}  \\
			
			\hline
		\end{tabular}
	\end{center}
	\caption{Monodular depth estimation results on KITTI dataset by the split of Eigen \etal~\cite{eigen}. K and CS refer to KITTI and Cityscapes datasets, respectively. As for supervision, `Depth' means the ground truth depth is used during training, `Stereo' means stereo image sequences with known baselines between two cameras are used during training, and `-' means no supervision is provided. The results are capped at 80m and 50m, respectively. As for error metrics Abs Rel, Seq Rel, RMSE and RMSE log, lower value is better; as for accuracy metrics $\delta<1.25$, $\delta<1.25^{2}$ and $\delta<1.25^{3}$, higher value is better.}
	\label{deptheval}
\end{table*}

As shown in Fig.~\ref{framework}, our framework includes 4 sub-networks. Both DepthNet and PoseMaskNet consist of encoding and decoding parts. The encoders are made up of 6 convolutional downsampling layers with stride 2, and decoders transform the extracted features into depth or masks with deconvolutional layers. Both depth and masks are predicted in 4 scales. In order to preserve both high-level and detailed information of the image, skip connections are used between encoders and decoders at corresponding resolutions~\cite{SfMLearner}. Meanwhile, the encoding part of PoseMaskNet is also followed by 2 fully-connected layers to regress Euler angles and translations of 6-DoF pose, respectively. The Encoder and discriminator follow the same architecture as the encoding part of DepthNet. The extracted feature from Encoder then passes through an average pooling layer to output a 128-channel vector. Batch normalization and ReLUs are adopted in each layer except for the output layers.

Our model is implemented by PyTorch~\cite{pytorch} on a single NVIDIA GTX 1080Ti GPU. All sub-networks are trained together in an self-supervised manner. During training, images are resized to 128$\times$416 and data augmentation (random rotation, zoom, color jitter) is applied to prevent overfitting. As suggested in WGAN~\cite{WGAN}, the stochastic gradient descent is used for the discriminator, and Adam~\cite{adam} optimizer with $\beta_{1}=0.9$, $\beta_{2}=0.99$ is used for all the other networks. The length of LSTM is set 15, and weighting factors $\lambda_{a}$, $\lambda_{s}$, $\lambda_{t}$, $\lambda_{g}$ are set 0.75, 0.1, 0.14 and 0.01, respectively. The training batch size is set 4 with a weight decay of $3\times10^{-4}$ for 100,000 iterations. The initial learning rate is set $10^{-4}$ and reduced by half for every 15,000 iterations. The network infers depth and pose at the speed of $18ms$ per frame during the test.

\subsection{Depth estimation}

We take the split of Eigen \etal~\cite{eigen} and use monocular images to train and test depth estimation. Ground truth depth is obtained by projecting sparse laser-scanned depth points into images, and depth predictions are interpolated to be the same size as ground truth for evaluation. In order to solve the scale ambiguity problem, the predicted depth is multiplied by a scaling factor to match the median with ground truth. Following the evaluation protocol in~\cite{benchmark}, both 50m and 80m thresholds of maximum depth are used for evaluation. As with previous methods, we also pre-train the network on Cityscapes dataset~\cite{cityscapes} and fine-tune on KITTI to test its adaptability across different environments.

We provide a comparison with related works which have depth supervision~\cite{eigen} or calibrated stereo images with known camera baseline for self-supervision. As shown in Table~\ref{deptheval}, our method outperforms all self-supervised methods and achieves comparable results with supervised ones. In particular, KITTI and Cityscapes datasets differ not only in scene contents but also in camera intrinsics. Results in the bottom rows of Table~\ref{deptheval} show that our method generalizes well in different environments. Since enhanced edges and details only take up a small proportion of depth maps, the improvement on depth accuracy is therefore limited.

\begin{figure*}
	\begin{center}
		\includegraphics[width=1.0\linewidth]{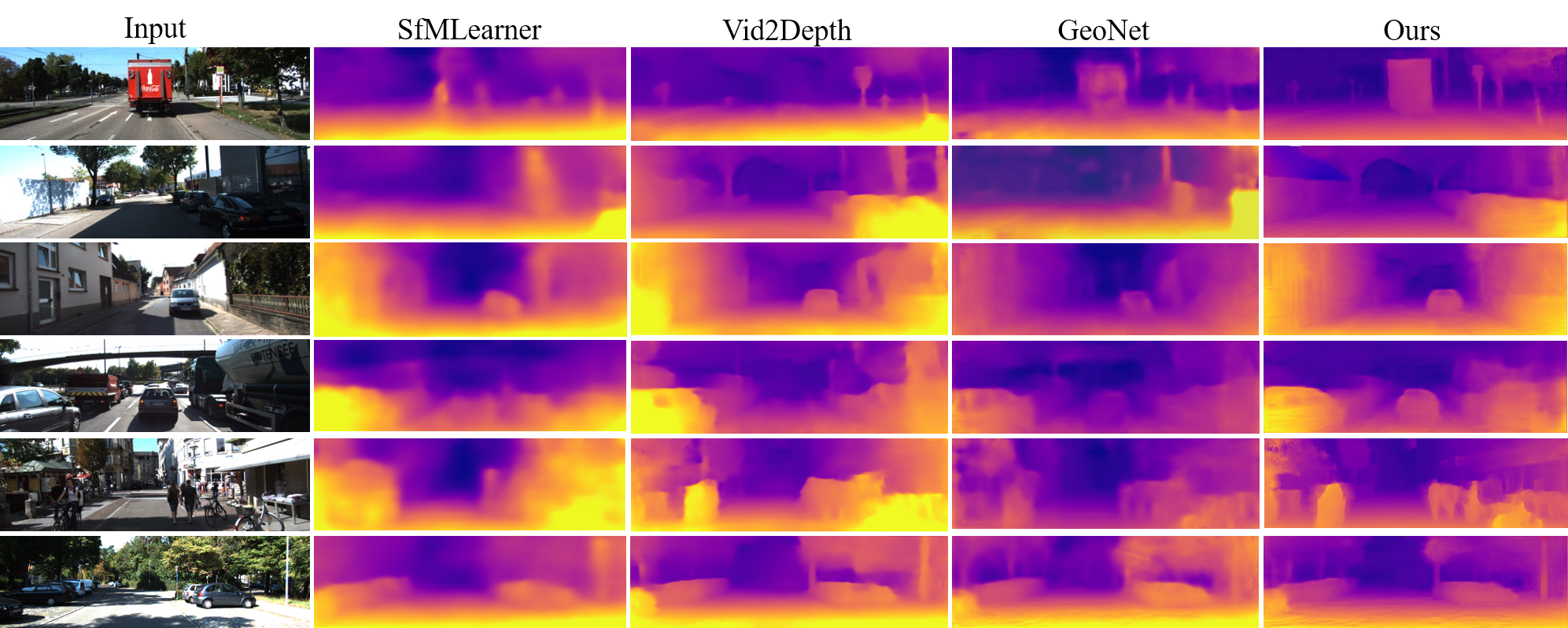}
	\end{center}
	\caption{Selected depth estimations from the test on KITTI dataset. Our method shows better prediction on detailed structures, low texture regions and shaded areas than the other self-supervised VO approaches. The estimated depth is clear in both close and distant areas.}
	\label{fig:depth}
\end{figure*}

Fig.~\ref{fig:depth} shows the qualitative examples of depth estimated by different methods. It can be seen that some methods have difficulty in recovering the depth of cars and mistake the depth of several objects. As the code provides frame-to-frame correspondence, our method produces clearer depth compared with single-view depth estimation approaches. Additionally, benefited from adversarial learning, the estimated depth preserves boundaries and thin structures, which is more accurate in details.

\subsection{Pose estimation}
In addition, we apply our method to KITTI odometry dataset for pose estimation. The dataset contains 11 driving scenes with ground truth poses. In order to make fair comparison, we follow the same train/test split as~\cite{GeoNet,SfMLearner} by using sequences 00-08 for training and 09-10 for test.

\begin{figure}
	\begin{center}
		\includegraphics[width=1\linewidth]{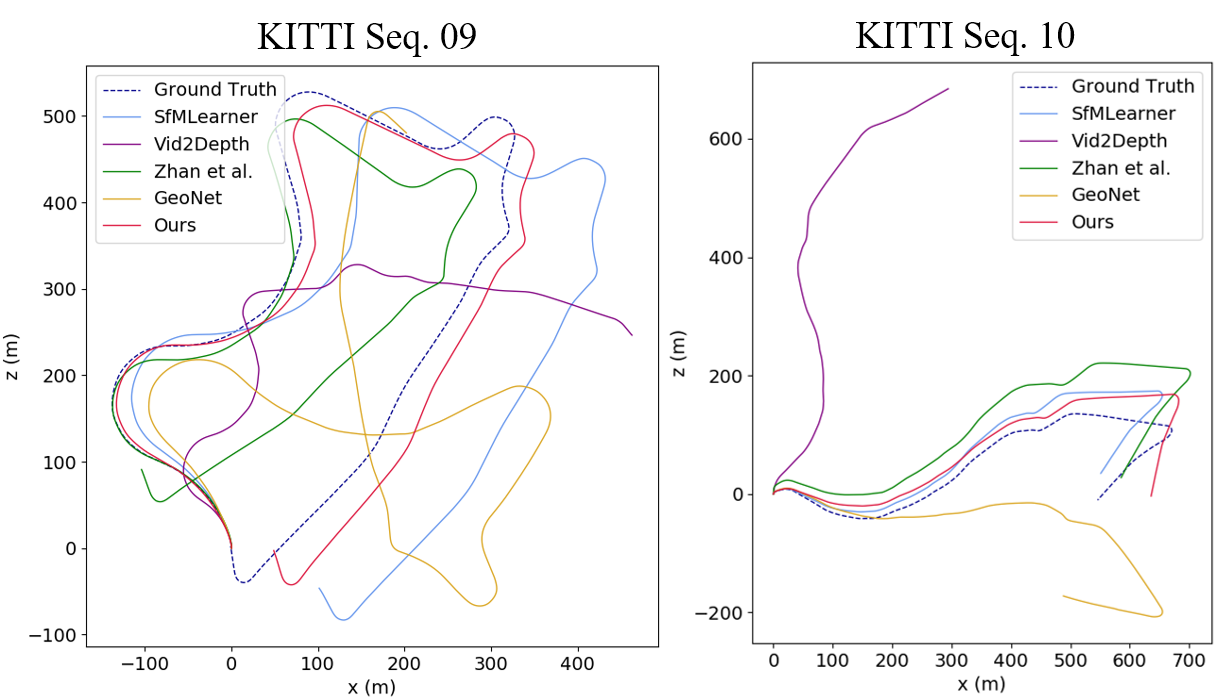}
	\end{center}
	\caption{Trajectories of different methods on KITTI dataset. Our method shows a better odometry in both rotation and translation.}
	\label{pose}
\end{figure}

The performance of pose estimation is evaluated using Absolute Trajectory Error (ATE) for both translation and rotation. Our method is compared with SfMLearner~\cite{SfMLearner}, GeoNet~\cite{GeoNet}, Vid2Dpeth~\cite{vid2depth}, Zhan \etal~\cite{deepvofeat} and ORB-SLAM, a representative framework in classic SLAM. ORB-SLAM (short) is emplemented by tracking module with local bundle adjustment, and ORB-SLAM (full) processes the entire sequence with loop closure and global bundle adjustment. Both versions of ORB-SLAM use a single scale map which is beneficial to an accurate trajectory with consistent scale. In order to solve the scale ambiguity problem in monocular VO, a scaling factor is used to align the trajectories with ground truth~\cite{deepvofeat}.

\begin{table}
	\footnotesize
	\begin{center}
		\begin{tabular}{|l|c|c|}
			\hline
			Method                                & Seq.09            & Seq.10          \\
			\hline
			ORB-SLAM~\cite{orb} (short)           & 0.064$\pm$0.141   & 0.064$\pm$0.130 \\
			ORB-SLAM~\cite{orb} (full)            & 0.014$\pm$0.008   & 0.012$\pm$0.011 \\
			SfMLearner~\cite{SfMLearner}          & 0.021$\pm$0.017   & 0.020$\pm$0.015 \\
			SfMLearner~\cite{SfMLearner} modified & 0.016$\pm$0.009   & 0.013$\pm$0.009 \\
			Zhan \etal~\cite{deepvofeat} & 0.013$\pm$0.009 & 0.013$\pm$0.008   \\
			Vid2Depth~\cite{vid2depth}            & 0.013$\pm$0.010   & 0.012$\pm$0.011 \\
			GeoNet~\cite{GeoNet}                  & 0.012$\pm$0.007   & 0.012$\pm$0.009 \\
			Ours                                  & \textbf{0.0030$\pm$0.0014} & \textbf{0.0029$\pm$0.0012} \\
			\hline
		\end{tabular}
	\end{center}
	\caption{Absolute Trajectory Error (ATE) on sequence 09 and 10 in KITTI odometry dataset. Our method outperforms all the other baselines by a large margin.}
	\label{ATE}
\end{table}

\begin{table*}
	\footnotesize
	\begin{center}
		\begin{tabular}{|l|c|c|c|c|c|c|c|c|c|}
			\hline
			Method             & Dataset & Cap & Abs Rel & Sq Rel & RMSE & RMSE log & $\delta<1.25$ & $\delta<1.25^{2}$ & $\delta<1.25^{3}$ \\
			\hline
			Baseline & K & 50m & 0.218 & 1.462 & 5.837 & 0.275 & 0.723 & 0.908 & 0.967 \\
			Baseline+code & K & 50m & 0.162 & 1.178 & 4.533 & 0.236 & 0.811 & 0.933 & 0.973 \\
			Baseline+code+GAN & K & 50m & 0.152 & 0.937 & 4.120 & 0.217 & 0.816 & 0.939 & 0.979 \\
			Baseline+code+LSTM & K & 50m & 0.148 & 0.939 & 4.271 & 0.217 & 0.816 & 0.941 & 0.977 \\
			Baseline+code+GAN+LSTM & K & 50m & 0.150 & 0.931 & 4.116 & 0.216 & 0.819 & 0.943 & 0.979 \\
			Baseline+code+GAN+LSTM+TC & K   & 50m  & \textbf{0.146} & \textbf{0.927} & \textbf{4.107} & \textbf{0.216} & \textbf{0.819} & \textbf{0.943} & \textbf{0.981}  \\
			\hline
		\end{tabular}
	\end{center}
	\caption{Ablation study on depth estimation for various versions of our method. Baseline denotes our framework without code, LSTM, discriminator (\ie GAN) and trajectory consistency (TC) loss.}
	\label{ablation_depth_table}
\end{table*}

\begin{figure*}
	\begin{center}
		\includegraphics[width=1\linewidth]{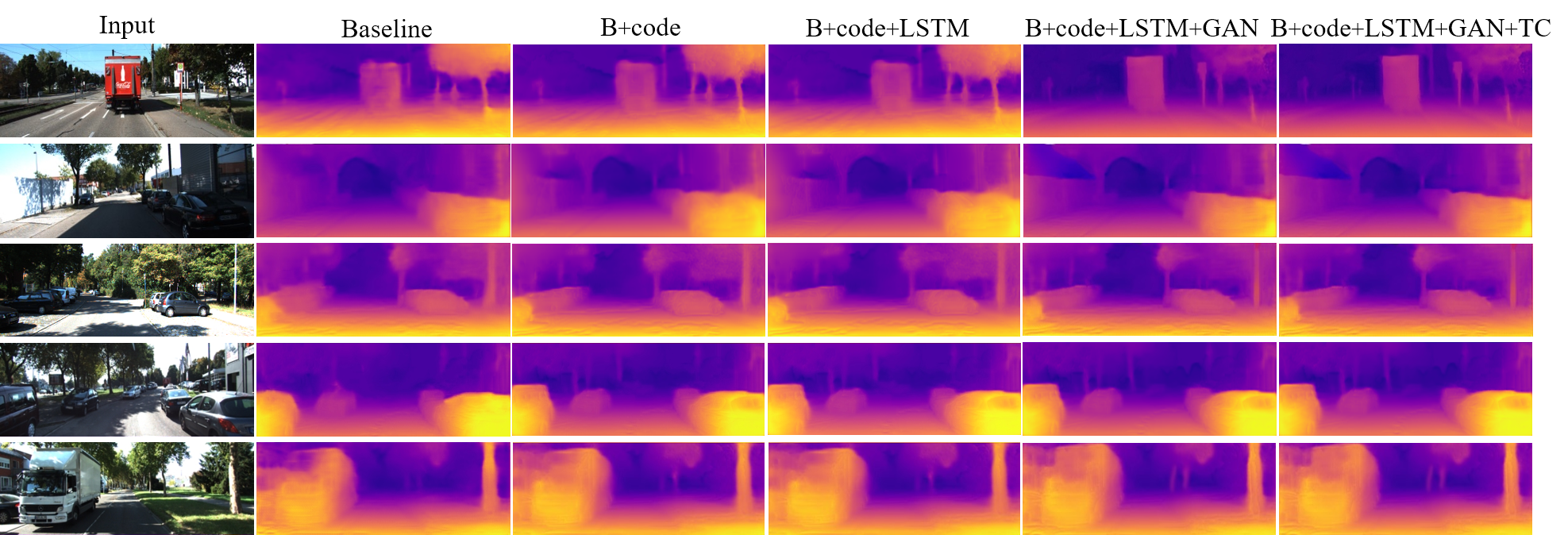}
	\end{center}
	\caption{Ablation study on depth estimation of our method. B denotes our baseline method, which is our framework without code, LSTM, discriminator (\ie GAN) and trajectory consistency (TC) loss.}
	\label{ablationdepth}
\end{figure*}

As shown in Table~\ref{ATE}, our method significantly outperforms all the other baselines, and trajectories of sequences 09-10 are plotted in Fig.~\ref{pose}. In addition, although only a limited number of frames can be processed by LSTM, our method still performs better than ORB-SLAM (full) without any need of global optimization (such as loop closure, bundle adjustment and re-localization)~\cite{orb}. This reveals that our method is able to produce accurate pose estimations by incorporating short-term correspondences and long-term dependences in odometry.

\subsection{Ablation studies}
In order to study the importance of each component, we perform ablation studies on various versions of our method. The baseline is our framework removing code, LSTM, trajectory consistency loss and discriminator. All the experiments are conducted on KITTI dataset and results are shown in Table~\ref{ablation_depth_table},~\ref{ablation_ATE} and Fig.~\ref{ablationdepth}.

As shown in Fig.~\ref{ablationdepth} (b), single view depth estimation is prone to be misled by the texture and color distributions in RGB images. The depth of poles is not recovered, and the depth of the sky is regarded the same as the white wall due to similar colors. In contrast, our method avoids these problems by taking additional information into account. The code encodes frame-to-frame correspondence which provides a significant improvement in depth estimation. The recovered depth is much sharper in contours and preserves tiny objects in both close and distant areas. In addition, adversarial learning gives the performance a further boost, and the temporal information actually improves depth.

\begin{table}
	\footnotesize
	\begin{center}
		\begin{tabular}{|l|c|c|}
			\hline
			Method                         & Seq.09            & Seq.10          \\
			\hline
			Baseline                        & 0.0072$\pm$0.0025   & 0.0070$\pm$0.0023 \\
			B+code                       & 0.0069$\pm$0.0021   & 0.0065$\pm$0.0020 \\
			B+code+GAN                    & 0.0064$\pm$0.0019   & 0.0062$\pm$0.0019 \\
			B+code+LSTM                    & 0.0045$\pm$0.0015   & 0.0043$\pm$0.0015 \\
			B+code+GAN+LSTM & 0.0036$\pm$0.0013   & 0.0036$\pm$0.0012 \\
			B+code+GAN+LSTM+TC & \textbf{0.0030$\pm$0.0014} & \textbf{0.0029$\pm$0.0012} \\
			\hline
		\end{tabular}
	\end{center}
	\caption{Ablation study on pose estimation for various versions of our method on KITTI sequence 09 and 10. B denotes baseline.}
	\label{ablation_ATE}
\end{table}

As for pose estimation in Table~\ref{ablation_ATE}, our baseline method performs much better than the other self-supervised VO approaches in literature (Table~\ref{ATE}). This may mainly because of the joint use of depth and image for pose estimation (Eq.~\eqref{equation_pose}). In addition, the accuracy is significantly improved by LSTM which incorporates historical information of multiple frames. The enforcement of trajectory consistency also brings about promising improvements in that it enforces geometric consistency among multiple pose estimations. Since depth is improved mainly on edges and details which takes up a small proportion, the accuracy gain is therefore limited. Yet the improved details are very important to RGBD matching for pose regression. Therefore, a slight increase in depth accuracy causes a big improvement in pose estimation.

\section{Conclusions}
We proposed an self-supervised VO framework that reduces accumulated errors over long sequence to achieve accurate pose and depth estimation. Benefited from spatial-temporal consistency among consecutive frames, the proposed framework incorporates historical information to reduce estimation errors in a self-supervised manner. In addition, we proposed to tackle VO as an self-supervised image generation task by means of a GAN paradigm. Our method outperforms both self-supervised and traditional VO baselines in literature, and ablation studies validate the effectiveness of each component of our framework.

In the future, we will extend our framework to unsupervised end-to-end SLAM. It is also worthwhile to investigate the code learned by our framework, which may help semantic segmentation, surface normal estimation and dense 3D reconstruction. In addition, developing an self-supervised online refinement technique to adaptively learn new environments on the fly is also an interesting issue of VO/SLAM and other 3D computer vision researches.

\textbf{Acknowledgments.}
The work is supported by the National Key Research and Development Program of China (2017YFB1002601) and National Natural Science Foundation of China (61632003, 61771026).

{\small
	\bibliographystyle{ieee_fullname}
	\bibliography{egbib}

\begin{thebibliography}{10}\itemsep=-1pt

\bibitem{GANdepth}
Filippo Aleotti, Fabio Tosi, Matteo Poggi, and Stefano Mattoccia.
\newblock {Generative Adversarial Networks for Unsupervised Monocular Depth
  Prediction}.
\newblock In {\em ECCV}, 2018.

\bibitem{WGAN}
Martin Arjovsky, Soumith Chintala, and L{\'e}on Bottou.
\newblock {Wasserstein Generative Adversarial Networks}.
\newblock In {\em ICML}, 2017.

\bibitem{deeperundemon}
V Babu, Anima Majumder, Kaushik Das, Swagat Kumar, et~al.
\newblock {A Deeper Insight into the UnDEMoN: Unsupervised Deep Network for
  Depth and Ego-Motion Estimation}.
\newblock {\em arXiv preprint arXiv:1809.00969}, 2018.

\bibitem{undemon}
V~Madhu Babu, Kaushik Das, Anima Majumdar, and Swagat Kumar.
\newblock {UnDEMoN: Unsupervised Deep Network for Depth and Ego-Motion
  Estimation}.
\newblock In {\em IROS}, 2018.

\bibitem{driven}
Dan Barnes, Will Maddern, Geoffrey Pascoe, and Ingmar Posner.
\newblock {Driven to Distraction: Self-Supervised Distractor Learning for
  Robust Monocular Visual Odometry in Urban Environments}.
\newblock In {\em ICRA}, 2018.

\bibitem{codeslam}
Michael Bloesch, Jan Czarnowski, Ronald Clark, Stefan Leutenegger, and Andrew~J
  Davison.
\newblock {CodeSLAM: Learning a Compact, Optimisable Representation for Dense
  Visual SLAM}.
\newblock In {\em CVPR}, 2018.

\bibitem{driving}
Chenyi Chen, Ari Seff, Alain Kornhauser, and Jianxiong Xiao.
\newblock {DeepDriving: Learning Affordance for Direct Perception in Autonomous
  Driving}.
\newblock In {\em ICCV}, 2015.

\bibitem{cityscapes}
Marius Cordts, Mohamed Omran, Sebastian Ramos, Timo Rehfeld, Markus Enzweiler,
  Rodrigo Benenson, Uwe Franke, Stefan Roth, and Bernt Schiele.
\newblock {The Cityscapes Dataset for Semantic Urban Scene Understanding}.
\newblock In {\em CVPR}, 2016.

\bibitem{flownet}
Alexey Dosovitskiy, Philipp Fischer, Eddy Ilg, Philip Hausser, Caner Hazirbas,
  Vladimir Golkov, Patrick Van Der~Smagt, Daniel Cremers, and Thomas Brox.
\newblock {FlowNet: Learning Optical Flow with Convolutional Networks}.
\newblock In {\em ICCV}, 2015.

\bibitem{eigen}
David Eigen, Christian Puhrsch, and Rob Fergus.
\newblock {Depth Map Prediction from a Single Image Using a Multi-Scale Deep
  Network}.
\newblock In {\em NIPS}, 2014.

\bibitem{DSO}
Jakob Engel, Vladlen Koltun, and Daniel Cremers.
\newblock {Direct Sparse Odometry}.
\newblock {\em IEEE Transactions on Pattern Analysis and Machine Intelligence},
  40(3):611--625, 2018.

\bibitem{LSD}
Jakob Engel, Thomas Sch{\"o}ps, and Daniel Cremers.
\newblock {LSD-SLAM: Large-Scale Direct Monocular SLAM}.
\newblock In {\em ECCV}, 2014.

\bibitem{classicopticalflow}
Gunnar Farnebäck.
\newblock {Two-Frame Motion Estimation Based on Polynomial Expansion}.
\newblock In {\em Scandinavian Conference on Image Analysis}, 2003.

\bibitem{robot}
Christian Forster, Simon Lynen, Laurent Kneip, and Davide Scaramuzza.
\newblock {Collaborative Monocular SLAM with Multiple Micro Aerial Vehicles}.
\newblock In {\em IROS}, 2013.

\bibitem{svo}
Christian Forster, Matia Pizzoli, and Davide Scaramuzza.
\newblock {SVO: Fast Semi-Direct Monocular Visual Odometry}.
\newblock In {\em ICRA}, 2014.

\bibitem{garg}
Ravi Garg, Vijay~Kumar BG, Gustavo Carneiro, and Ian Reid.
\newblock {Unsupervised CNN for Single View Depth Estimation: Geometry to the
  Rescue}.
\newblock In {\em ECCV}, 2016.

\bibitem{benchmark}
Cl{\'e}ment Godard, Oisin Mac~Aodha, and Gabriel~J Brostow.
\newblock {Unsupervised Monocular Depth Estimation with Left-Right
  Consistency}.
\newblock In {\em CVPR}, 2017.

\bibitem{GAN}
Ian Goodfellow, Jean Pouget-Abadie, Mehdi Mirza, Bing Xu, David Warde-Farley,
  Sherjil Ozair, Aaron Courville, and Yoshua Bengio.
\newblock {Generative Adversarial Nets}.
\newblock In {\em NIPS}, 2014.

\bibitem{mapnet}
Joao~F Henriques and Andrea Vedaldi.
\newblock {MapNet: An Allocentric Spatial Memory for Mapping Environments}.
\newblock In {\em CVPR}, 2018.

\bibitem{LSTM}
Sepp Hochreiter and J{\"u}rgen Schmidhuber.
\newblock {Long Short-Term Memory}.
\newblock {\em Neural Computation}, 9(8):1735--1780, 1997.

\bibitem{pix2pix}
Phillip Isola, Junyan Zhu, Tinghui Zhou, and Alexei~A Efros.
\newblock {Image-to-Image Translation with Conditional Adversarial Networks}.
\newblock In {\em CVPR}, 2017.

\bibitem{CTC}
Ganesh Iyer, J Krishna~Murthy, Gunshi Gupta, Madhava Krishna, and Liam Paull.
\newblock {Geometric Consistency for Self-Supervised End-to-End Visual
  Odometry}.
\newblock In {\em CVPR Workshops}, 2018.

\bibitem{RGBDreg}
Christian Kerl, Jurgen Sturm, and Daniel Cremers.
\newblock {Dense visual SLAM for RGB-D cameras}.
\newblock In {\em IROS}, 2014.

\bibitem{adam}
Diederik~P Kingma and Jimmy Ba.
\newblock {Adam: A method for Stochastic Optimization}.
\newblock In {\em ICLR}, 2015.

\bibitem{PTAM}
Georg Klein and David Murray.
\newblock {Parallel Tracking and Mapping on a Camera Phone}.
\newblock In {\em ISMAR}, 2009.

\bibitem{undeepvo}
Ruihao Li, Sen Wang, Zhiqiang Long, and Dongbing Gu.
\newblock {UndeepVO: Monocular Visual Odometry through Unsupervised Deep
  Learning}.
\newblock In {\em ICRA}, 2018.

\bibitem{liu}
Fayao Liu, Chunhua Shen, Guosheng Lin, and Ian Reid.
\newblock {Learning Depth from Single Monocular Images Using Deep Convolutional
  Neural Fields}.
\newblock {\em IEEE Transactions on Pattern Analysis and Machine Intelligence},
  38(10):2024--2039, 2016.

\bibitem{vid2depth}
Reza Mahjourian, Martin Wicke, and Anelia Angelova.
\newblock {Unsupervised Learning of Depth and Ego-Motion from Monocular Video
  Using 3D Geometric Constraints}.
\newblock In {\em CVPR}, 2018.

\bibitem{orb}
Raul Mur-Artal, Jose Maria~Martinez Montiel, and Juan~D Tardos.
\newblock {ORB-SLAM: A Versatile and Accurate Monocular SLAM System}.
\newblock {\em IEEE Transactions on Robotics}, 31(5):1147--1163, 2015.

\bibitem{VR}
Richard~A Newcombe, Shahram Izadi, Otmar Hilliges, David Molyneaux, David Kim,
  Andrew~J Davison, Pushmeet Kohi, Jamie Shotton, Steve Hodges, and Andrew
  Fitzgibbon.
\newblock {KinectFusion: Real-Time Dense Surface Mapping and Tracking}.
\newblock In {\em ISMAR}, 2011.

\bibitem{RGBDreg2}
Jaesik Park, Qian~Yi Zhou, and Vladlen Koltun.
\newblock {Colored Point Cloud Registration Revisited}.
\newblock In {\em ICCV}, 2017.

\bibitem{pytorch}
Adam Paszke, Sam Gross, Soumith Chintala, and Gregory Chanan.
\newblock {PyTorch}.
\newblock \url{https://github.com/pytorch/pytorch}, 2017.

\bibitem{demon}
Benjamin Ummenhofer, Huizhong Zhou, Jonas Uhrig, Nikolaus Mayer, Eddy Ilg,
  Alexey Dosovitskiy, and Thomas Brox.
\newblock {DeMoN: Depth and Motion Network for Learning Monocular Stereo}.
\newblock In {\em CVPR}, 2017.

\bibitem{deepvo}
Sen Wang, Ronald Clark, Hongkai Wen, and Niki Trigoni.
\newblock {DeepVO: Towards End-to-End Visual Odometry with Deep Recurrent
  Convolutional Neural Networks}.
\newblock In {\em ICRA}, 2017.

\bibitem{ssim}
Zhou Wang, Alan~C Bovik, Hamid~R Sheikh, Eero~P Simoncelli, et~al.
\newblock {Image Quality Assessment: from Error Visibility to Structural
  Similarity}.
\newblock {\em IEEE Transactions on Image Processing}, 13(4):600--612, 2004.

\bibitem{guided}
Fei Xue, Qiuyuan Wang, Xin Wang, Wei Dong, Junqiu Wang, and Hongbin Zha.
\newblock {Guided Feature Selection for Deep Visual Odometry}.
\newblock In {\em ACCV}, 2018.

\bibitem{beyond}
Fei Xue, Xin Wang, Shunkai Li, Qiuyuan Wang, Junqiu Wang, and Hongbin Zha.
\newblock {Beyond Tracking: Selecting Memory and Refining Poses for Deep Visual
  Odometry}.
\newblock In {\em CVPR}, 2019.

\bibitem{DVSO}
Nan Yang, Rui Wang, Jorg Stuckler, and Daniel Cremers.
\newblock {Deep Virtual Stereo Odometry: Leveraging Deep Depth Prediction for
  Monocular Direct Sparse Odometry}.
\newblock In {\em ECCV}, 2018.

\bibitem{GeoNet}
Zhichao Yin and Jianping Shi.
\newblock {GeoNet: Unsupervised Learning of Dense Depth, Optical Flow and
  Camera Pose}.
\newblock In {\em CVPR}, 2018.

\bibitem{deepvofeat}
Huangying Zhan, Ravi Garg, Chamara Saroj~Weerasekera, Kejie Li, Harsh Agarwal,
  and Ian Reid.
\newblock {Unsupervised Learning of Monocular Depth Estimation and Visual
  Odometry with Deep Feature Reconstruction}.
\newblock In {\em CVPR}, 2018.

\bibitem{deeptam}
Huizhong Zhou, Benjamin Ummenhofer, and Thomas Brox.
\newblock {DeepTAM: Deep Tracking and Mapping}.
\newblock In {\em ECCV}, 2018.

\bibitem{SfMLearner}
Tinghui Zhou, Matthew Brown, Noah Snavely, and David~G Lowe.
\newblock {Unsupervised Learning of Depth and Ego-Motion from Video}.
\newblock In {\em CVPR}, 2017.

\bibitem{cyclegan}
Jun~Yan Zhu, Taesung Park, Phillip Isola, and Alexei~A. Efros.
\newblock {Unpaired Image-to-Image Translation Using Cycle-Consistent
  Adversarial Networks}.
\newblock In {\em ICCV}, 2017.

\end{thebibliography}
}

\end{document}